# Virtual Sensor Based Fault Detection and Classification on a Plasma Etch Reactor

*Donald A. Sofge, NeuroDyne Inc.*

**Abstract**

The SEMATECH sponsored J-88-E project teaming Texas Instruments with NeuroDyne (et al) focused on Fault Detection and Classification (FDC) on a Lam 9600 aluminum plasma etch reactor, used in the process of semiconductor fabrication. Fault classification was accomplished by implementing a series of virtual sensor models which used data from real sensors (Lam Station sensors, Optical Emission Spectroscopy, and RF Monitoring) to predict recipe setpoints and wafer state characteristics. Fault detection and classification were performed by comparing predicted recipe and wafer state values with expected values. Models utilized include linear PLS, Polynomial PLS, and Neural Network PLS. Prediction of recipe setpoints based upon sensor data provides a capability for cross-checking that the machine is maintaining the desired setpoints. Wafer state characteristics such as Line Width Reduction and Remaining Oxide were estimated on-line using these same process sensors (Lam, OES, RFM). Wafer-to-wafer measurement of these characteristics in a production setting (where typically this information may be only sparsely available, if at all, after batch processing runs with numerous wafers have been completed) would provide important information to the operator that the process is or is not producing wafers within acceptable bounds of product quality. Production yield is increased, and correspondingly per unit cost is reduced, by providing the operator with the opportunity to adjust the process or machine before etching more wafers.

## 1.0 Background

The ability to sense and adapt to varying material characteristics and process conditions over a large range of operating conditions is critical to the affordable, high volume manufacture of IC electronic devices. In a flexible manufacturing environment this is highly dependent upon the accurate development and subsequent adaptation of models which simulate process, wafer, and equipment relationships and with feedback from in-situ sensors are used to predict process trends and develop control strategies. Virtual sensor models are shown to be capable of predicting machine states and wafer state properties such as line width and oxide loss based upon process sensor data (machine state sensors, Optical Emission Spectroscopy (OES), RF Monitoring (RFM)). Improvements in sensor based feedback and control that remove uncertainty in plasma etching will have a major impact in semiconductor manufacturing and integrated circuit fabrication. As plasma etch is a key step in many semiconductor fabrication processes, improvements in plasma etch using virtual sensor based models provide a crucial link to intelligent process monitoring and sensor-based control in the multi-billion dollar semiconductor manufacturing industry.

A key automation problem in the semiconductor manufacturing area is the efficient high-yield fabrication of semiconductor circuits. Plasma etching, a dry etching technique that usually follows the growth or deposition of thin films, is the key process by which desired circuits are patterned on a semiconductor wafer. As pattern geometries become more intricate in the submicron range, etching processes become more complex. In a typical etching process, a mixture of different halogen-containing gases are introduced in a vacuum etching chamber. The plasma is generated in the reactor by a high-frequency RF source. The desired goals of the etching process are controlling performance parameters such as the etch rate, the selectivity of etch for process endpoint control, the anisotropy for feature size control, and minimum defect generation. A number of process factors influence all of these parameters of interest: flow rate, power density of the RF source, pressure, chemistry, purity of the environment, substrate bias, and electrode configuration. Because the plasma process is highly nonlinear, controllability of the desired parameters is considered intractable. Typical semiconductor manufacturers use a trial and error procedure to realize a repeatable fabrication process that has acceptable yield. This is expensive in time and material. Better modelling, instrumentation and control techniques that remove this uncertainty in etching will have a major impact in semiconductor manufacturing and integrated circuit fabrication.

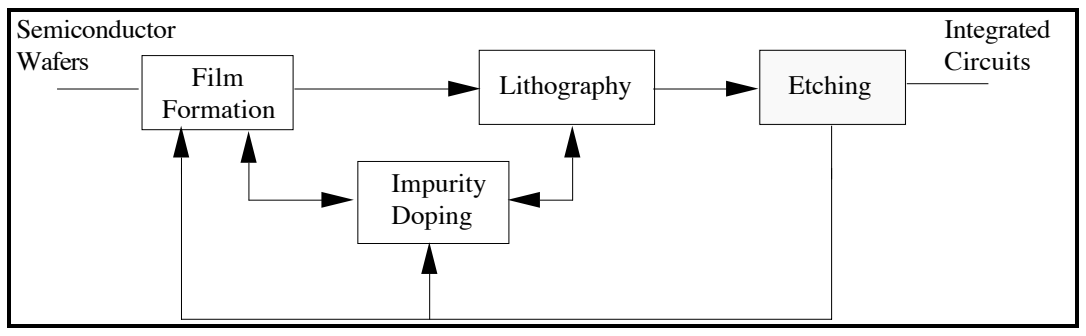

**Figure 1.** Sequence of major process steps in silicon integrated circuit fabrication

## 1.1 Plasma Etching

While chemical or wet etching was the norm in previous generations of IC processing, it is limited to feature sizes of 1-2 mm and aspect ratios of the devices no better than 1:1. Since the state of the art of today's and the next generation high-density ICs will rely on submicron features as low as 0.1 mm, dry etching techniques based on plasma etching has become the dominant etching process. Plasma etching, such as those based on electron cyclotron resonant (ECR) sources and reactive ion etch (RIE), allow etching of fine lines and features without loss of definition. In this form of etching, a plasma, comprising ions, free radicals and neutral species, is formed above a masked surface by adding large energy doses to a gas at low pressures. This is commonly accomplished by electrical discharges in gases at milliTorr pressures generating high kinetic energy plasma that impinge on the non-masked portions of the semiconductor substrate. By using gases that react with the substrate, the etching can be made more effective. Typically, halogen-containing gases, such as $CF_4$ and $CHF_3$, are used together with other gases such as oxygen. A horizontal parallel plate radial flow type plasma reactor that is used for plasma etching (or deposition by changing the chemistry) is shown in **Figure 2**.

The etching process is described and specified by various parameters which may include:

- Line Width
- Oxide Loss
- Etch rate
- Selectivity: relative etch rate of different materials
- Anisotropy: ratio of vertical to horizontal etch rates
- Uniformity: refers to variations in etching rate among runs, among wafers, or across a wafer
- Defect density on the wafer: these arise due to particulate matter generated during the etching process; expressed as number of point defects/$cm^2$

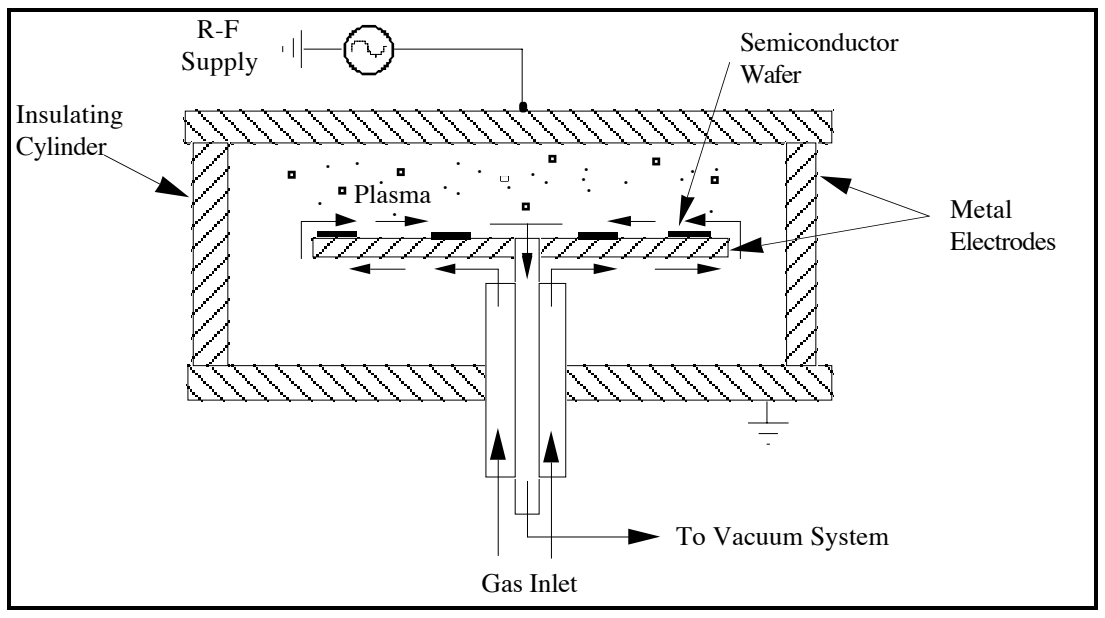

**Figure 2**. A parallel plate radial flow plasma reactor

## 1.2 Virtual Sensor Modelling

Plasma etch processes in use today have been made reproducible by insuring that the process parameters are set according to a recipe. The parameter settings are selected to provide a broad process window. The assumption has been if all of the process parameters that affect the process are set correctly, the process environment and the product produced will be reproducible. Although substantial progress has been made using this approach, significant problems remain unresolved. Of particular importance is the need to be able to quickly detect a condition in the machine or the process environment that will have adverse effects on the product.

The process model representation is shown in **Figure 3**. The process or plasma etch chamber is shown in the center of the figure. The function **f** maps from recipe setpoint parameters to chamber states, while the function **g** maps from the chamber states to wafer states. Also shown is the inverse mapping f(-1), which is the virtual sensor mapping from the chamber state sensors back to recipe setpoints. The functions **f(-1)** and **g** are implemented as virtual sensor models.

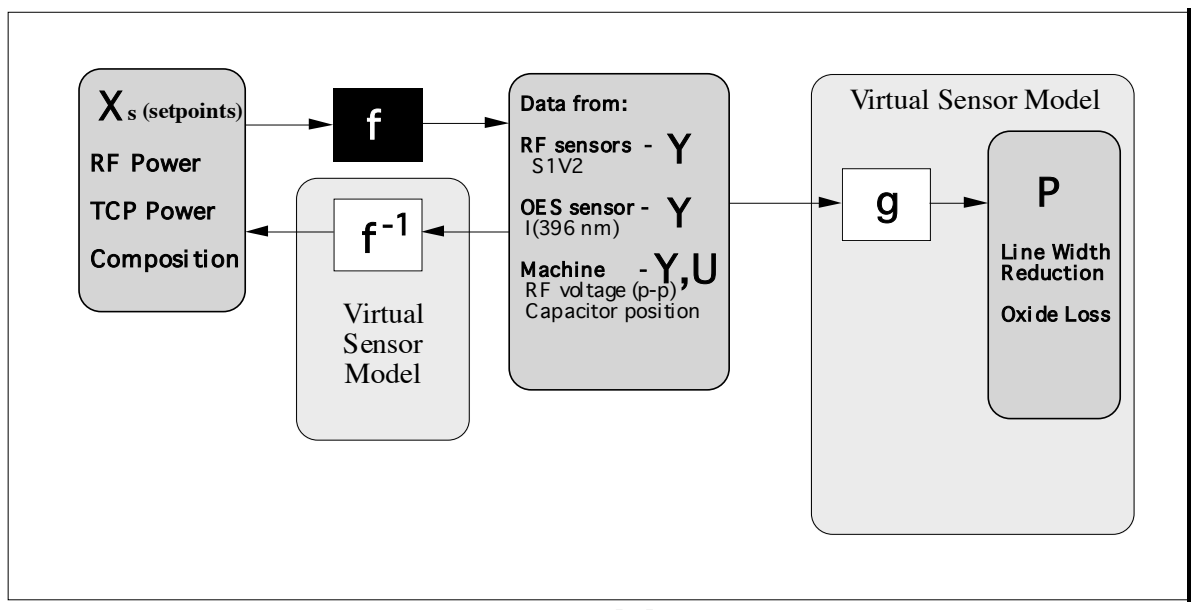

**Figure 3.** Process Model Representation

The use of sensor measurements for estimating setpoints and wafer states is based on the premise that the large number of signals from machine sensors, from optical emission spectroscopy (OES) sensors and from RFM sensors is rich in information about the "true" state of the plasma etch processing. If one can sort out the time series of hundreds of signals from these sensors and look at them in the appropriate way, it is possible to predict important information about the process and product quality. Wafer state characteristics which normally can only be measured by meticulous testing after processing is complete could be estimated routinely in real-time in a production environment, through use of Virtual Wafer State sensors. Recipe setpoints could be verified even in the presence of sensor drift and possible sensor failure through use of virtual setpoint sensors.

| | Machine | OES | RFM |
|---|---|---|---|
| Recipe Parameters | $f^{-1}$ model | $f^{-1}$ model | $f^{-1}$ model |
| Line Width Reduction | g model | g model | g model |
| Oxide Loss (Etch Rate) | g model | g model | g model |

**Figure 4.** Multiple Virtual Sensors Provide Orthogonal Estimates of Process and Wafer States

Furthermore, if the actual sensors providing the data to the virtual sensors are completely independent from one another (such as OES and RFM), then the use of multiple virtual sensors using orthogonal (independent) measurements could be used to provide redundant estimates of wafer states and setpoints as shown in **Figure 4**. This use of redundancy would further insulate the process against sensor drift, and if virtual sensor estimates agree, would provide further support that the virtual sensor estimates are reliable indicators of true process and wafer states.

## 2.0 Implementation

The purpose of this program was to demonstrate advanced fault detection and classification for plasma etching on a Lam 9600 metal etcher. Process sensors including an Optical Emission Spectroscopy system and RF Monitoring sensors were added to monitor conditions in the plasma etch chamber. Computer interfaces for capturing the data and storing it to disk were created. Virtual sensor models were implemented using multivariate statistical methods including PLS and PCR (and others described below), as well as techniques which combined neural networks with statistical methods. Fault classification was accomplished by implementing a series of virtual sensor models which use data from real sensors (such as Machine State, OES, and RFM) to predict recipe setpoints ($f^{-1}$) and wafer state characteristics (g). Prediction of recipe setpoints based upon sensor data provides a capability for cross-checking that the machine is maintaining the desired setpoints, and may indicate sensor drift or failure if virtual sensors agree with one another but disagree with recipe setpoint values. Wafer state characteristics such as Line Width Reduction and Oxide Loss may be estimated on-line (g model) using these same process sensors (Machine, OES, RFM). Wafer-to-wafer measurement of these characteristics in a production setting (where typically this information may be only sparsely available, if at all, after batch processing runs with numerous wafers have been completed) would provide important information to the operator that the process is or is not producing wafers within acceptable bounds of product quality (e.g. LWR). If the g model virtual sensors, for example, reported a bad Oxide Loss after a given wafer had been processed, the operator would have the opportunity to adjust the process or machine before etching more wafers.

### 2.1 Design Of Experiments (DOEs)

Since one of the goals was to model the plasma etch process for a wide variety of process conditions and across a wide range of setpoints (rather than for just a single recipe), an experimental design was created to attempt to span the range of setpoints of interests. This resulted in two sets of wafer experiments, referred to as DOEs (design of experiments).

The DOEs (Exp30 and Exp32) were based on a 5 level central composite design centered around recipe 44, with 70 wafers in each DOE. Each of the DOEs spans 3 lots of wafers, or together they span 6 lots. Exp32 was designed to replicate Exp30 and provide temporal robustness to the models. Of these 70 wafers, 35 were designated for training the models, 23 for cross-validation, and 12 for testing.

### 2.2 Data Pretreatment

Raw sensor measurements from wafer processing are recorded every few seconds (exact sampling rates depend upon the specific sensor system), sometimes at irregular intervals, and generally the sampling of these signals is not coordinated with the sampling times for other sensors connected to the same machine. Each of the sensors suites described provide dozens of raw sensor measurements at each sampling interval. The time history of these signals from processing a single wafer provides the sensor data record of the etch. Since the etch times for individual wafers vary, the length of these data records also varies. Since these data records are quite voluminous and cumbersome to analyze in this format, for purposes of virtual sensor modelling in this effort it was decided to reduce the data through pretreatment. Data pretreatment is covered in detail elsewhere in this report, so only the types of pretreatment used for building the $f^{-1}$ and g models needs to be mentioned here. OES data was first pretreated by reducing 2042 spectral lines into 40. Next, the time series records for sensor measurements were reduced a to set of vectors of signal metrics (means, std, etc.) for each wafer processed. This pretreament not only greatly simplified the modelling, but also enhanced model precision through precalculation of a number of important metrics which turned out to be very useful for prediction.

### 2.2.1 Data Separated by Etch Region

One aspect of the TAS pretreatment is that it separates the sensor data by etch region (**Figure 5**). In a previous phase of this project we analyzed the predictive capability of virtual sensor models using data individually from one of the three main etch regions (Al, TiN, Ox), as well as the usefulness of combining the data from all etch regions. It was found that there is great variation in the predictive capability of models by etch region (some parameters modelled better from TiN region data, some better with Ox region data). In addition it was found that certain combinations of sensor type and etch region provided better data for certain models (e.g. OES based models using Ox region data provided the best $f^{-1}$ models, while RFM based models benefited most from TiN region data for all predictions). Combining sensor data from multiple etch regions, based on the premise that there might be a significant amount of complementary data present at different stages of the etch, yielded worse not better predictions. From this result it was decided to focus in this phase of the project on use of data from etch regions individually (to not combine them).

| | Machine | OES | RFM |
|---|---|---|---|
| Al Etch Region | **24 TAS metrics** {e.g. Endpoint_A Al avg, tcp_match_tuning_cap Al sigma} | **126 TAS metrics** {e.g. 388 Al S1 avg, 388 Al S2 avg, 388 Al S3 avg} | **140 TAS metrics** {e.g. S1V1 Al avg, S1V1 Al sigma, S1V2 Al avg, S1V2 Al sigma} |
| TiN Etch Region | **60 TAS metrics** {e.g. Endpoint_A TiN adjR$^2$, Endpoint_A TiN avg, Endpoint_A TiN sigma, Endpoint_A TiN max, Endpoint_A TiN min} | **252 TAS metrics** {e.g. 388 TiN S1 adjR$^2$, 388 TiN S2 adjR$^2$, 388 TiN S3 adjR$^2$, 388 TiN S1 max, 388 TiN S2 max, 388 TiN S3 max} | **280 TAS metrics** {e.g. S1V2 TiN adjR$^2$, S1V2 TiN ssr, S1V2 TiN sigma, S1V2 TiN max} |
| Ox Etch Region | **24 TAS metrics** {e.g. Endpoint_A Ox avg, tcp_match_tuning_cap Ox sigma} | **126 TAS metrics** {e.g. 388 Ox S1 avg, 388 Ox S2 avg, 388 Ox S3 avg} | **140 TAS metrics** {e.g. S1V1 Ox avg, S1V1 Ox sigma, S1V2 Ox avg, S1V2 Ox sigma} |

**Figure 5.** Sensor Data Metrics are Divided by Etch Region

## 2.3 Modelling Techniques Examined

A wide variety of modelling techniques for implementation of the virtual sensor models were analyzed. These included the following:

- Multidimensional Linear Regression (MLR)
- Principal Component Regression (PCR)
- Linear Partial Least Squares (PLS)
- Polynomial Regression
- Polynomial Partial Least Squares (PolyPLS)
- Neural Network Partial Least Squares (NNPLS)

It was determined that MLR often failed on this type of data due to the collinearity of many of the signals (they are highly correlated), so this technique was abandoned. Polynomial regression was far less accurate in most cases than the other techniques tried, so this was dropped too. PCR and PLS generally yielded similar results (there are linear, polynomial, and neural network versions of PCR as well as for PLS), but PLS was generally slightly better than PCR. Therefore for this stage of the project it was decided to focus only on the use of PLS techniques, considering only Linear PLS, Polynomial PLS, and Neural Network PLS models.

In addition to verifying that wafer state parameters and process setpoints *can* in fact be modelled using process sensor data, we sought to determine which modelling techniques would be most suitable for this task, which etch region(s) provided the richest source(s) of information for prediction, how accurate and how robust would these models be. It was decided that we would model the 7 recipe parameters included in the DOE (the $f^{-1}$ models), and two wafer state parameters (LWR and Oxide Loss). Models were developed separately for each sensor data set to achieve the redundancy described, and based upon data from individual etch regions. Furthermore, the question of which of the three PLS techniques (Linear PLS, Polynomial PLS, and Neural Network PLS) would provide the most robust, accurate models needed to be addressed.

Of the three modelling techniques used (linear, polynomial, and neural network based PLS), the least accurate and least robust of these techniques was clearly polynomial PLS. The best models for a given sensor and etch region generally came from either neural network based PLS or linear PLS, with the neural network based PLS often edging out the linear PLS in terms of prediction accuracy. Often, however, for the same training and prediction data, the difference between these two models was slight. More importantly, however, the linear PLS technique proved to be more robust. While the neural network based PLS often gave the highest prediction accuracy, it sometimes also gave the worst (on data from a different etch region/sensor combination). Due to this somewhat unpredictable behavior on the part of neural network based PLS, and the marginal advantage it offered in terms of prediction accuracy, it was decided to focus on the analysis of linear PLS results for the remainder of the effort.

## 3.0 Modelling Results

### 3.1 $f^{-1}$ Model Results, Sensor Measurements to Setpoints

The purpose of the $f^{-1}$ virtual sensor model is to use process state sensor to predict recipe setpoint values. This is to provide a way of cross-checking the effective setpoint parameters according to plasma chamber dynamics with the desired setpoints as specified by the current recipe. If there is a mismatch between what the setpoints are and what the $f^{-1}$ virtual sensor models are predicting, then it is possible that the process has drifted from setpoint and needs to be corrected. It can also indicate that the sensors and/or actuators regulating setpoints may be in error due to miscalibration, drift or malfunction.

The recipe parameters modelled for $f^{-1}$ were:

- Pressure (mTorr)
- Top Power (watts)
- Rfbot (watts)
- $BCl_3$ (sccm)
- $Cl_2$ (sccm)
- $Cl_2/BCl_3$
- Total Flow (sccm)

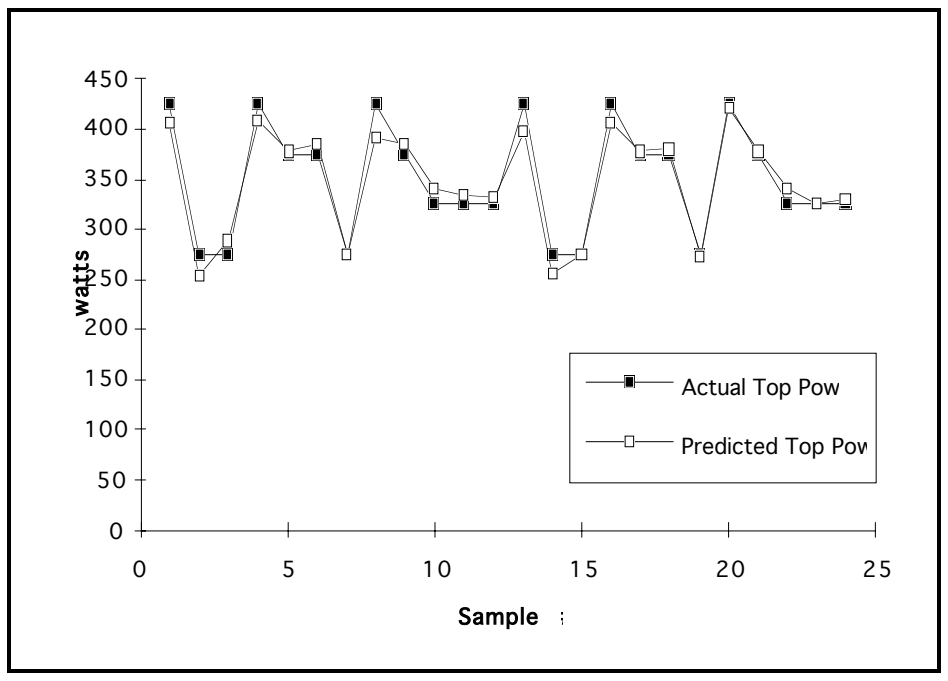

**Figure 6.** Linear PLS Model of Top Power from RFM Sensors, Ox Region

As shown in **Figures 6** and **7**, it was possible to get fairly accurate predictive models for the power parameters, by carefully selecting sensor type and etch region which resulted in the best model(s). Note that these are pure predictions based on data which were not included in model building. It should be noted that there were many models passed over because they did not perform so well. There appeared to be little consistency from one DOE to the next on which etch regions and which sensors would result in the best models, though some areas and sensors did appear to give consistently poor results.

As shown in **Figure 8**, the best models gave fairly accurate results (usually less than 5%) for predicting the RF parameters, and significantly poorer results (20-30% error) in predicting the gas flow parameters.

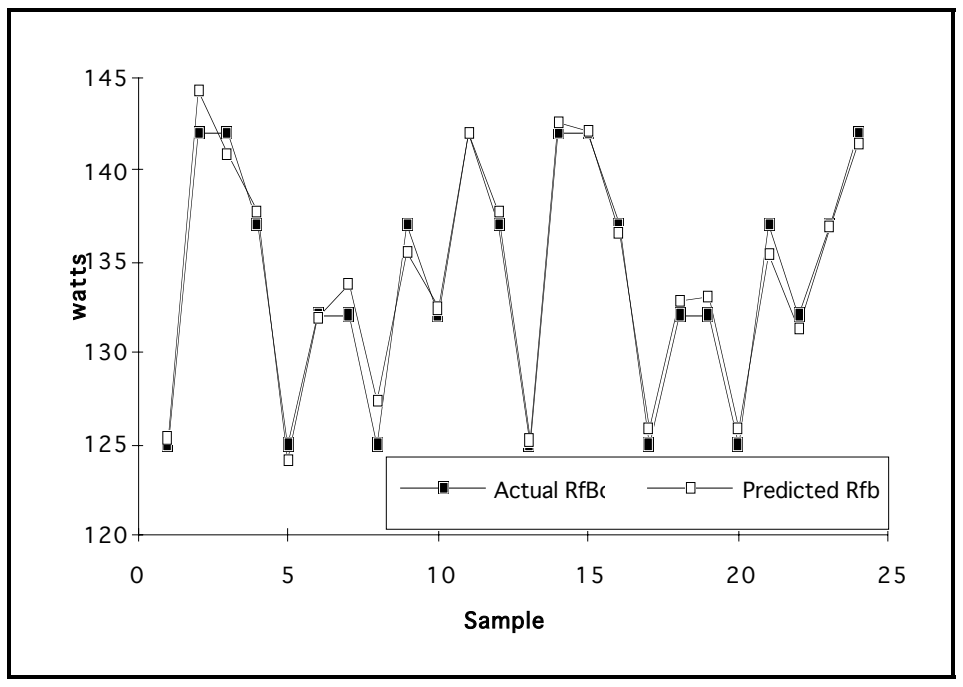

**Figure 7.** Linear PLS Model of Rfbot from machine sensors, Ox Region

| | Machine | OES | RFM |
|---|---|---|---|
| Pressure | 1 mTorr | 2 mTorr | 2 mTorr |
| Top Power | 11 watts | 31 watts | 9 watts |
| Rfbot | 2 watts | 10 watts | 2 watts |
| $BCl_3$ | 6 sccm | 6 sccm | 9 sccm |
| $Cl_2$ | 7 sccm | 7 sccm | 7 sccm |
| $Cl_2/BCl_3$ | 0.1 | 0.1 | 0.1 |
| Total Flow | 10 sccm | 12 sccm | 14 sccm |

**Figure 8.** Summary of $f^{-1}$ Model Predictive Capabilities (RMS Prediction Error)

### 3.2 Model g Results, Sensor Measurements to Wafer States

#### 3.2.1 Line Width Reduction (LWR)

Analysis of Electrical LWR data clearly indicated that there was a significant variation of LWR based upon position on the wafer. Electrical line width measurements were taken post-etch for the 32 die positions on each wafer. Since there are no die location specific variables in the process sensors (although there is some OES sensor sensitivity to stripes of die locations, depending upon the orientations of the OES fiber optic sensors), it was necessary to build a separate PLS model for each die position. (This is functionally the equivalent of having a multiple Y-block PLS model which has a separate prediction for each die).

No pre-etch measurements were taken, so all LWR measurements were based upon an assumed incoming line width of 0.5 microns. In fact it was later found that there was a significant variation of incoming line width (described elsewhere in this report) of about 0.02 microns. Given this limitation in the accuracy of the post-etch LWR data ((0.5 - measurement) +/-0.02 microns), then we could only expect to model LWR to about 0.02 microns.

Comparison of results from using Neural Network based PLS models to Linear PLS models illustrates a common result found in this study: that while the NNPLS models may have the lowest average prediction error (NNPLS OES Ox models have the highest prediction accuracy), the NNPLS technique may also result in some of the worst models (NNPLS RFM Al models). The Linear PLS models are almost as accurate as the NNPLS models, but don't seem to result in really poor models as often.

This point is made clearer by looking at the prediction accuracies of the individual die models for LWR. As shown in **Figure 9**, there is a clear dependence on model accuracy and die position (which may be an indication of LWR reproducibility at different locations on a wafer). It is also obvious that the NNPLS models are more accurate than the Linear PLS models.

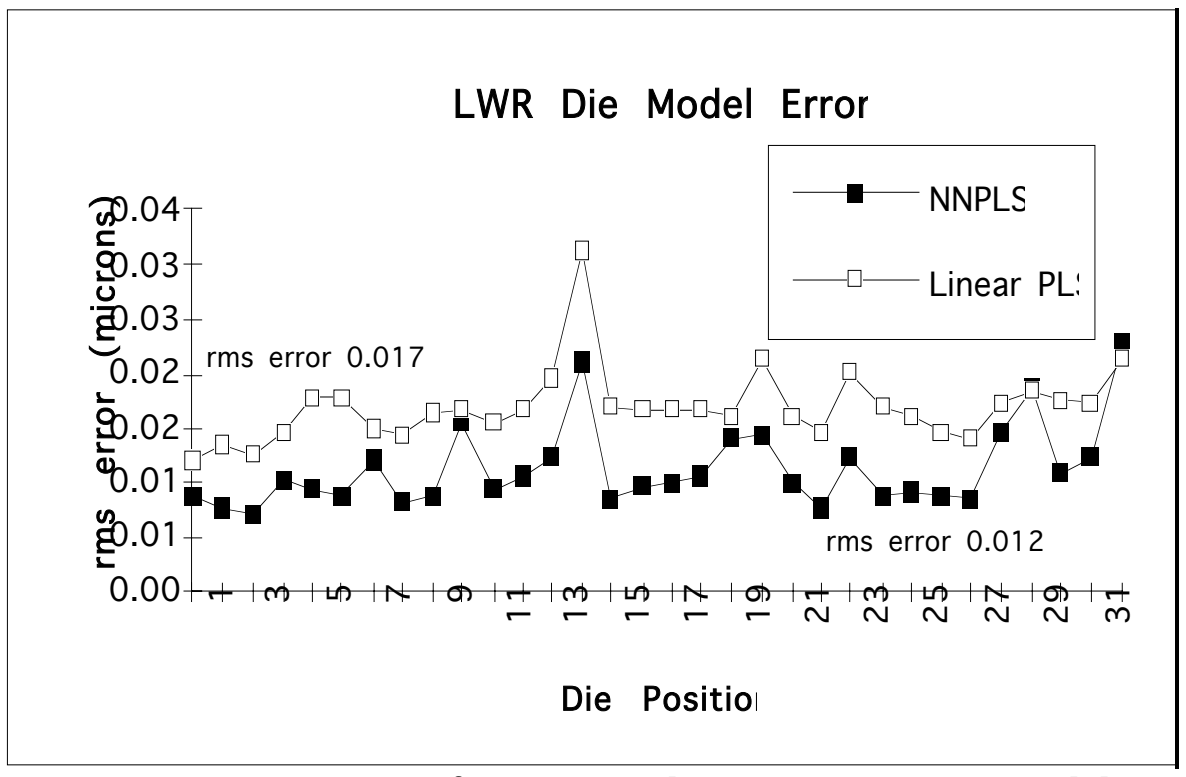

**Figure 9.** Comparison of NNPLS and Linear PLS LWR Models Showing Model Prediction Errors for Each Die

3.2.2 Oxide Loss

As with LWR measurements, there were 32 oxide loss measurements taken per wafer, with a dependence of oxide loss measurement upon die position, so it was necessary to generate a separate model for each die position. Looking at the oxide loss predictions from this sensor and etch region for Die #32 (**Figure 10**), we can see that the model was able to track oxide loss pretty well. Note in particular the large deviation in wafer #8 (for the die #32 position) of the test set. The Linear PLS model was quite successful in predicting this deviation strictly based upon the machine state data.

Since there was no "nominal" oxide loss available for these DOE data, we calculated the average oxide loss across all dies across all wafers. Note that this average doesn't represent a normal or expected value, since the DOEs spanned a wide range of recipes and processing conditions, but merely represent a statistical mean. It was found that we could predict oxide loss from each of the process sensors using Linear PLS models to within about 6-7% of the range of the training data.

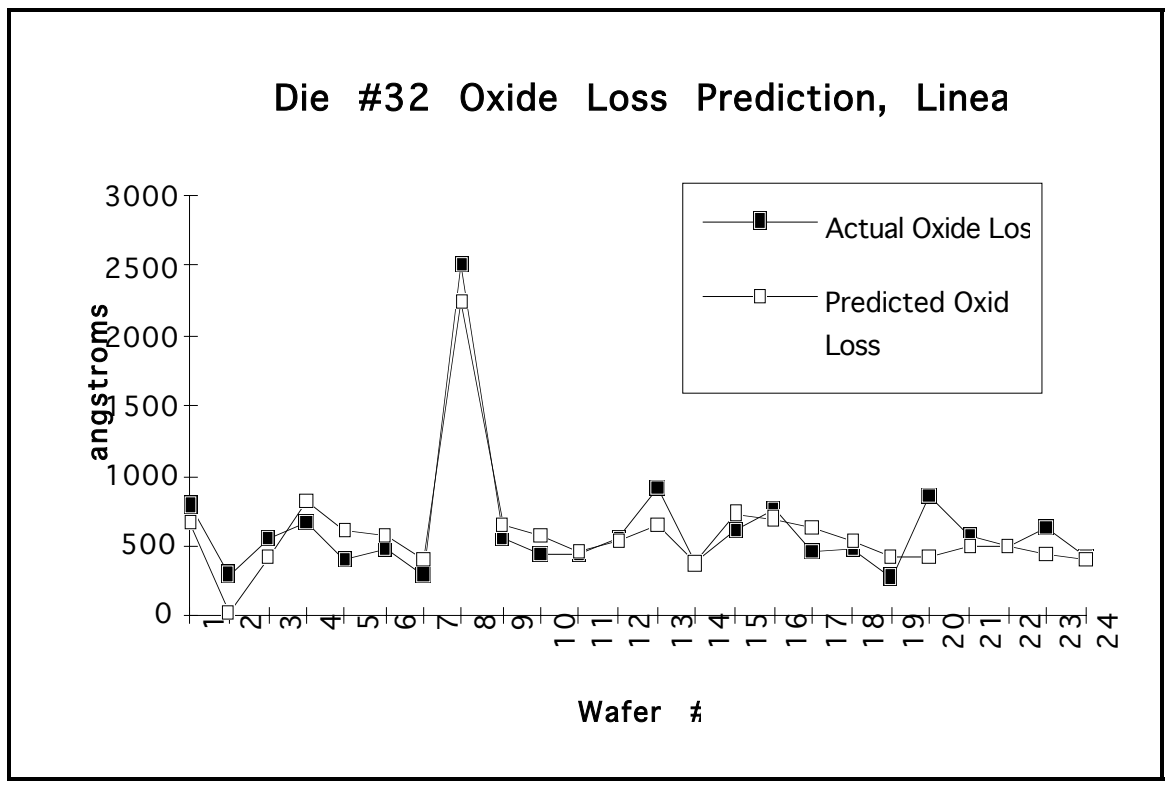

**Figure 10.** Actual Model Predictions for Die #32 from Machine Sensor Data, TiN Etch Region

## 4.0 Conclusions

Accurate predictions of Wafer State Characteristics and Recipe Parameters can be achieved using virtual sensor models with Machine State, OES, and RFM Sensor Data. Each of these sensors provides sufficient information for accurate predictions. No one sensor is consistently better than the others. Use of multiple redundant virtual sensors for each quantity being sensed is necessary, since predictions from a given sensor and etch region may not be consistently reliable. Virtual sensor models are made robust by including in their training sets sensor data which spans multiple lots over an extended period of time. Models built upon data from a single lot of wafers processed at a particular point in time are not likely to be valid outside of that lot and point in time.

Three PLS modelling techniques were investigated in this effort: Linear PLS, Polynomial PLS, and Neural Network PLS (NNPLS). NNPLS and Linear PLS provided the best results (minimum rms prediction error). Although NNPLS often provides slightly better models than Linear PLS, NNPLS also produces the worst models in many cases. If only one modelling technique were to be used, we would recommend that it be Linear PLS. Linear PLS generally provides models comparable to the best NNPLS models (if sometimes slightly less accurate), while resulting in fewer really poor models than the other techniques tried.

## 5.0 Future Research

Accurate predictions of wafer state characteristics and recipe parameters can be achieved using virtual sensor models with machine state, optical emission spectroscopy (OES), and RF monitoring (RFM) sensor data. In many cases each of these sensors provide sufficient information for accurate predictions. However, significant levels of noise still exist in many of the sensor signals which corrupt models, and many of the sensor signals are highly correlated. Our results have suggested that by selectively reducing the number of variables used for modelling, we can improve the overall robustness of our models and get consistently better predictions. By separating over a thousand different variables (108 machine state variables, 504 OES variables, and 560 RFM variables) according to etch region, we found that we were able to improve the accuracy of our predictive models. Also, contrary to intuition, combining or fusing sensor data from two or more sensors into a single model generally resulted in worse, not better, models. An analysis of the variables suggests that significant noise exists in many of the sensor streams, and that the addition of more sensor data has the potential to corrupt models rather than improve them. Also, it was found through multivariate statistical analysis that many of these variables (for all three sensor systems mentioned) were highly correlated.

This suggests that models may be made more accurate and much more robust if we can intelligently select which combinations of variables from the sensor data which will result in the best models, while eliminating those which will corrupt the models. Variable selection is needed into order to select a set or subset of sensor lines which contribute to models with the best predictive accuracy and robustness. Since there are literally hundreds of variables to comb through, we propose using Genetic Algorithms to select variables, build and test models, and to evolve a set of variables which yield models with better predictive capabilities and which are consistently more accurate.